\documentclass[sigconf,authorversion]{acmart}

\usepackage{booktabs} 
\usepackage{amsmath}
\usepackage{algorithm}
\usepackage[noend]{algpseudocode}
\usepackage{amssymb}

\copyrightyear{2018} 
\acmYear{2018} 
\setcopyright{rightsretained}
\acmConference[GECCO '18]{Genetic and Evolutionary Computation Conference}{July 15--19, 2018}{Kyoto, Japan}
\acmBooktitle{GECCO '18: Genetic and Evolutionary Computation Conference, July 15--19, 2018, Kyoto, Japan}
\acmPrice{15.00}
\acmDOI{10.1145/3205455.3205602}
\acmISBN{978-1-4503-5618-3/18/07}

\DeclareMathOperator*{\argmax}{\arg\max}

\begin{document}

\title[Discovering the Elite Hypervolume by Leveraging Interspecies Correlation]{Discovering the Elite Hypervolume \\by Leveraging Interspecies Correlation}

\author{Vassilis Vassiliades}
\affiliation{Inria, CNRS, Universit\'{e} de Lorraine, F-54000 Nancy, France}
\email{vassilis.vassiliades@inria.fr}

\author{Jean-Baptiste Mouret}
\affiliation{Inria, CNRS, Universit\'{e} de Lorraine, F-54000 Nancy, France}
\email{jean-baptiste.mouret@inria.fr}

\renewcommand{\shortauthors}{Vassiliades and Mouret}

\begin{abstract}
	Evolution has produced an astonishing diversity of species, each filling a different niche. Algorithms like MAP-Elites mimic this divergent evolutionary process to find a set of behaviorally diverse but high-performing solutions, called the elites. Our key insight is that species in nature often share a surprisingly large part of their genome, in spite of occupying very different niches; similarly, the elites are likely to be concentrated in a specific ``elite hypervolume'' whose shape is defined by their common features. In this paper, we first introduce the elite hypervolume concept and propose two metrics to characterize it: the genotypic spread and the genotypic similarity. We then introduce a new variation operator, called ``directional variation'', that exploits interspecies (or inter-elites) correlations to accelerate the MAP-Elites algorithm. We demonstrate the effectiveness of this operator in three problems (a toy function, a redundant robotic arm, and a hexapod robot).
\end{abstract}




\maketitle

\section{Introduction}
The astonishing diversity and elegance of life forms has long been an inspiration for creative algorithms that attempt to mimic the evolutionary process. Nevertheless, current evolutionary algorithms primarily view evolution as an optimization process \cite{stanley_lehman_book}, that is, they aim at performance, not diversity. It is therefore no wonder that most experiments in evolutionary computation do not show an explosion of diverse and surprising designs, but show instead a convergence to a single, rarely surprising ``solution''~\cite{stanley_lehman_book}. 

\begin{figure}[t]
	\includegraphics[width=0.7\columnwidth]{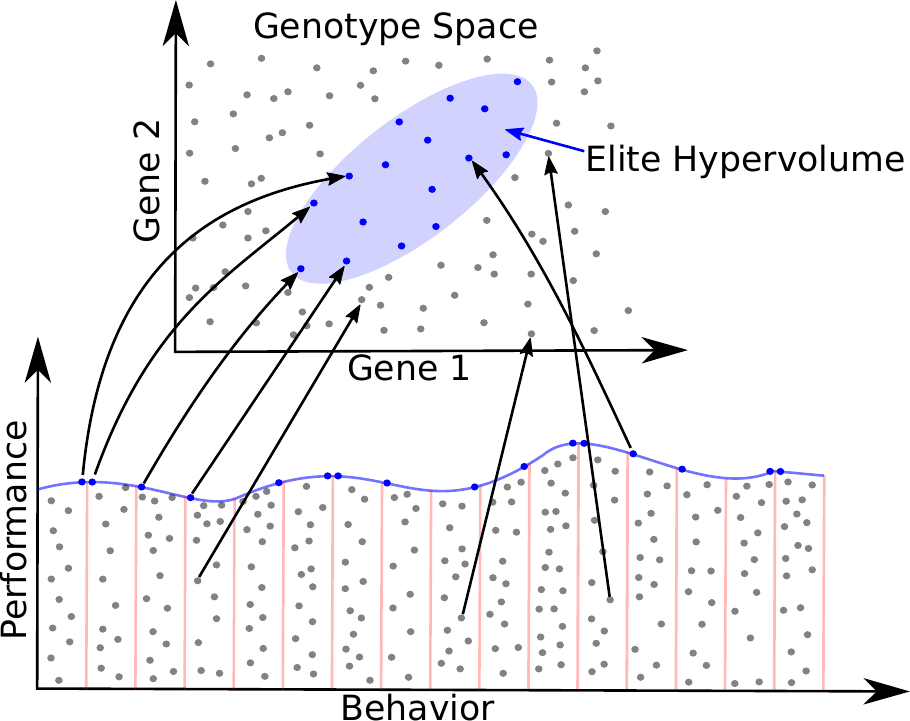}
	\caption{Illumination/quality diversity algorithms search for the highest-performing solution in each behavioral niche. These solutions are likely to be concentrated in a particular, elite hypervolume because neighboring high-performing solutions often have similar genotypic features.}
	\label{fig_conceptual}
\end{figure}

This view of artificial evolution has recently been challenged by a new family of algorithms that focus more on diversification than on optimization. This does not mean that performance --- fitness --- does not play any role: inside an ecological niche, individuals compete and optimize their fitness; but two species in two different niches are not competing directly. These algorithms are called ``illumination algorithms''~\cite{mouret2015illuminating}, because they ``illuminate the search space'', or ``quality diversity algorithms''~\cite{pugh2016quality}, because they search for a set of diverse, but high-performing solutions.

Current illumination algorithms essentially differ in the way they define niches and in how they bias the selection of individuals to reproduce \cite{cully2017quality,pugh2016quality}. We are here interested in the Multi-dimensional Archive of Phenotypic Elites (MAP-Elites) algorithm \cite{mouret2015illuminating,cully2015robots,vassiliades2017cvt}, because it produces high-quality results, while being conceptually simple and straightforward to implement. MAP-Elites explicitly divides the behavior space in niches using a regular grid \cite{mouret2015illuminating} or a centroidal Voronoi tessellation (CVT) \cite{vassiliades2017cvt} and each niche stores the best individual found so far, called the elite. MAP-Elites was successfully used to create high-performing behavioral repertoires for robots \cite{cully2015robots, duarte2016evorbc, vassiliades2017cvt, kume2017map, chatzilygeroudis2018rte}, design airfoils \cite{gaier2017gecco} and soft robots \cite{mouret2015illuminating}, evolve images that ``fool" deep neural networks~\cite{nguyen2015deep}, ``innovation engines" able to generate images that resemble natural objects~\cite{nguyen2016understanding}, and 3D-printable objects by leveraging feedback from neural networks trained on 2D images~\cite{lehman2016iccc}.

While being effective at generating a diverse set of high-performing solutions, MAP-Elites requires numerous fitness evaluations to do so. For instance, a few million evaluations are typically used when evolving behavioral repertoires for robots \cite{cully2015robots,chatzilygeroudis2018rte}. The objective of this paper is to propose an updated MAP-Elites that uses fewer evaluations for similar or better results.

Our main insight is that the high-performing solutions in different niches often share many traits, even when they look very different. In nature, a tiny worm like C. Elegans uses neurons and cells that are similar to those used by humans, insects, and all the other animal life forms. Similarly, all the bird species have a beak, two wings, a heart, and two lungs, but they occupy widely diverse niches, from the sea to tropical forests. Modern genomic analyses confirm this idea: species that occupy very different niches often share a surprisingly large part of their genome \cite{bejerano2004ultraconserved, pontius2007initial}. For example, fruit flies and humans share about 60 percent of their genes \cite{adams2000genome}.

If we translate the concept that ``high-performing species have many things in common'' to evolutionary computation, we conclude that all the elites of the search space, as found by MAP-Elites, are likely to be concentrated in a sub-part of the genotypic space (Fig. \ref{fig_conceptual}).
We propose to call this sub-part of the genotypic space the ``elite hypervolume''. Describing this sub-part would correspond to writing the ``recipe'' for high-performing solutions. For instance, all the high-speed walking controllers might need to use the same high-frequency oscillator, in spite of very different gait patterns. 
This idea might be counter-intuitive at first, because we would expect that a well spread set of behaviors would correspond to a well spread set of genotypes, but this is the case only when there exists a linear mapping between genotypes and behaviors. In the more general case, many genotypes can lead to the same behavior (e.g., there are many ways of not moving for a walking robot), and we should expect that competition between solutions will make it likely that survivors share some traits (e.g., a good balancing controller is useful for any gait).

The first objective of the present paper is to empirically show that elites are often concentrated in a small elite hypervolume whose shape reflects the common features of high-performing solutions. The second objective is to introduce a variation operator that exploits the correlations between the elites, that is, the ``interspecies'' similarities that define the elite hypervolume.

\section{Background}
\subsection{Illumination / Quality Diversity}

Illumination algorithms originated in the field of evolutionary robotics with the purpose of encouraging diversity in a space known as the behavior space~\cite{lehman2008exploiting}. This space describes the possible behaviors of individuals over their lifetimes: for example, a point in this space, i.e., a behavior descriptor, could be the final positions of simulated robots whose controllers are evolved~\cite{lehman2008exploiting}. In contrast, the genotype space is the space in which the evolutionary algorithm operates (e.g., a space of bit strings) and the phenotype space encodes the possible controllers (e.g., neural networks) that are derived from the genotype space.

Current illumination algorithms define niches and bias selection in different ways \cite{cully2017quality,pugh2016quality,lehman2011evolving}. The two main algorithms are currently MAP-Elites \cite{mouret2015illuminating}, which defines niches with a grid and selects uniformly among the elites, and Novelty Search with Local Competition \cite{lehman2011evolving,pugh2016quality}, which defines niches using a neighborhood based on a behavioral distance and selects using a multi-objective ranking between density in behavior space and performance. Depending on the task, selection biases can be introduced in all the algorithms. For instance, MAP-Elites can be modified to bias selection towards sparse regions~\cite{pugh2016quality, cully2017quality}, which can be facilitated using a tree structure~\cite{smith2016rapid}. 

To our knowledge, there is no explicit consideration for variation operators in the current algorithms: they use the variation operators that were designed for previous work with objective-based evolutionary algorithms (e.g., real-valued genetic algorithms (GAs)~\cite{agrawal1995sbx} or the NEAT operators for neural networks~\cite{stanley2002evolving}). 

\subsection{The CVT-MAP-Elites Algorithm} \label{sec_cvt}

\begin{algorithm}
	\caption{CVT-MAP-Elites algorithm}\label{algo:cvt_map_elites}
	\begin{algorithmic}[1]
		\Procedure{CVT-MAP-Elites}{$k$}
		\State $\mathcal{C} \longleftarrow$ \textsc{CVT}($k$) \Comment{Run CVT and get the centroids}
		\State $(\mathcal{X}, \mathcal{P}) \longleftarrow$ create\_empty\_archive($k$)
		\For {$i=1\to G$} \Comment{\emph{Initialization: $G$ random $\mathbf{x}$}}
		\State $\mathbf{x} = $ random\_individual()
		\State \textsc{add\_to\_archive}($\mathbf{x}, \mathcal{X}, \mathcal{P}$)
		\EndFor
		\For {$i=1\to I$} \Comment{\emph{Main loop, $I$ iterations}}
		\State $\mathbf{x} = $ selection($\mathcal{X}$)
		\State $\mathbf{x}' = $ variation($\mathbf{x}$)
		\State \textsc{add\_to\_archive}($\mathbf{x}', \mathcal{X}, \mathcal{P}$)
		\EndFor
		\State \textbf{return} archive ($\mathcal{X}, \mathcal{P}$)
		\EndProcedure
		
		\Procedure{add\_to\_archive}{$\mathbf{x}, \mathcal{X}, \mathcal{P}$}
		\State $(p,\mathbf{b}) \longleftarrow $ evaluate($\mathbf{x}$)
		\State $c \longleftarrow $ get\_index\_of\_closest\_centroid($\mathbf{b}, \mathcal{C}$)
		\If {$\mathcal{P}(c) = null$ or $\mathcal{P}(c) < p$}
		\State $\mathcal{P}(c) \longleftarrow p$, $\mathcal{X}(c) \longleftarrow \mathbf{x}$
		\EndIf
		\EndProcedure
	\end{algorithmic}
\end{algorithm}
We here use the CVT-MAP-Elites algorithm with uniform selection~\cite{vassiliades2017cvt, vassiliades2017unbounded}, which generalizes MAP-Elites to arbitrary dimensions and provides explicit control over the desired number of niches.

CVT-Map-Elites first partitions the behavior space into $k$ well-spread centroids (niches) using a CVT (Alg.~\ref{algo:cvt_map_elites}, line 2). It then creates an empty archive with capacity $k$ ($\mathcal{X}$ and $\mathcal{P}$ store the genotypes and performances, respectively). At the first generation, the algorithm samples a set of random genotypes (line 5) and evaluates them by recording their performance and behavior descriptor (line 13); it calculates the centroid closest to each behavior descriptor (line 14) and stores the individual in the archive only if the corresponding region is empty or has a less fit solution (lines 15,16). The main loop of the algorithm corresponds to selecting a random parent (line 8), varying the parent to create the offspring (line 9) and attempting to insert it in the archive as above. Note that there is no \textit{specific} strategy for variation, which we address in this paper.

\subsection{Exploiting Correlations in Evolutionary Algorithms}
\label{sec:review-cross}

In evolution strategies (ES), correlations between variables can be exploited by allowing each population member to maintain a multivariate Gaussian distribution (in the form of different mutation strengths and rotation angles)~\cite{schwefel1987collective}. Modern variants of ES (such as \cite{hansen2001completely} or \cite{wierstra2008natural}) and other estimation of distribution algorithms (EDAs)~\cite{larranaga2001estimation, bouter2017rvgomea}, exploit such correlations by building probabilistic models (that act as search distributions) from which they sample the next population. EDAs have been augmented with niching mechanisms to address multimodal optimization problems~\cite{shir2010adaptive, preuss2010cmaes, ahrari2017multimodal, maree2017niching};
however, they have never been used for illumination\footnote{A combination of ES and techniques from illumination algorithms have been proposed recently~\cite{conti2017improving}, but not for the purpose of illumination.}.

In GAs, commonalities between solutions can be exploited by the recombination operator. In real-coded GAs, parent-centric operators~\cite{deb2001self, deb2002computationally}, i.e., ones that create solutions near the parent with more probability, can be more beneficial than mean-centric ones (e.g., \cite{eshelman1993blx}), especially when the population has not surrounded the optimum~\cite{jain2011parent}. One of the most successful parent-centric operators is the simulated binary crossover (SBX)~\cite{agrawal1995sbx}, which creates two offspring from two randomly selected parents using a polynomial distribution. By spreading the offspring in proportion to the spread of the parents, SBX endows GAs with self-adaptive properties similar to ES~\cite{deb2001self}. A variant of SBX produces the offspring along a line that joins two parents~\cite{deb2006}, thus, being able to exploit linear correlations between them. To our knowledge, there has not been any study about exploiting correlations with a crossover operator when niching is performed either in phenotype or in behavior space.

\section{Tasks} \label{sec_tasks}

We perform our experiments in the following tasks, where $\mathbf{x} \in [0,1]^n$ is the genotype, $\mathbf{y}$ is the phenotype, and the genotype-phenotype map is a linear scaling to the range described below.

\paragraph{Schwefel's Function 1.2}
This is a classic function used when benchmarking optimization algorithms~\cite{schwefel1993evolution}. The objective is to maximize $f(\mathbf{y}) = -\sum_{i=1}^{n} \big( \sum_{j=1}^{i} y_j \big)^2$.
We use a 100-dimensional genotype space ($\mathbf{y} \in [-5,5]^{100}$) and set the behavior descriptor to be the first 2 phenotypic dimensions ($\mathbf{b(y)} \in [-5,5]^2$). 

\paragraph{Arm Repertoire}
The purpose of this experiment is to create a repertoire of joint angles for a redundant robotic arm for which the resulting end effector positions cover its reachable space~\cite{cully2015robots}. 
After convergence, each filled niche will contain a solution to its corresponding inverse kinematics (IK) problem, i.e., a joint configuration that takes the end effector inside the region; the goal of the illumination algorithm is, thus, to collect solutions for thousands of IK problems (one per region) in a single run. We use a 12-degree of freedom (DOF) arm in 2D space, where each joint is a revolute one (no joint limit), and we find the end effector position using the forward kinematics equations:
\begin{equation*}
\mathbf{b(y)} = 
\left[ {\begin{array}{c}
	l_1 \, cos(y_1) + l_2 \, cos(y_1 + y_2) + \dots + l_n \, cos(y_1 + \dots + y_n) \\
	l_1 \, sin(y_1) + l_2 \, sin(y_1 + y_2) + \dots + l_n \, sin(y_1 + \dots + y_n) \\
	\end{array} } \right]
\end{equation*}
where $n = 12$, each link length $l_i=1/n$, $y_i$ is a joint angle (thus, the phenotype space is the joint space, $\mathbf{y} \in [-\pi,\pi]^{12}$), and $\mathbf{b(y)} \in [-1,1]^2$ is the behavior descriptor.
The objective is to maximize the negative variance of the joint angles (straighter arms have higher performance)~\cite{cully2015robots}:
$f(\mathbf{y}) = -\frac{1}{n}\sum_{i=1}^{n} (y_i - \mu) \, \mbox{,} \quad \mbox{where} \, \mu = \sum_{i=1}^{n} y_i$.

\paragraph{Hexapod Locomotion}

The final experiment is the hexapod locomotion task of~\cite{cully2015robots}\footnote{We implemented this task using the Dynamic Animation and Robotics Toolkit (https://github.com/dartsim/dart).}. The objective is to maximize the distance covered by the hexapod robot in 5 seconds. The controller is an open-loop oscillator that actuates each motor by a periodic signal of frequency $1Hz$ parameterized by the amplitude, its phase shift and its duty cycle (i.e., the fraction of each period that the joint angle is positive). Each leg has 3 joints, however, only the movement of the first 2 is defined in the genotype (as the control signal of the third motor of each leg is the opposite of the second one)~\cite{cully2015robots}. This results in a 36D genotype space ($\mathbf{y} \in [0,1]^{36}$), as there are 6 parameters for each of the 6 legs of the robot. The behavior descriptor $\mathbf{b(y)} \in [0,1]^6$ is defined as the proportion of time each leg is in contact with the ground~\cite{cully2015robots} ($dt=15ms$, thus, the number of simulation steps $T = 5sec / 15ms = 333$).

\section{The Elite Hypervolume}
\subsection{Definition}

The elite hypervolume\footnote{Note how this is different from the concept of an ``$n$-dimensional hypervolume'' which was proposed in ecology (see~\cite{blonder2014n} and references therein) as a definition of a species niche in a multi-dimensional space of resources; in CVT-MAP-Elites, such a hypervolume would correspond to a single region in behavior space.}, $H$, is the subset of the $n$-dimensional real-valued genotype space, $X$, $H \subset X$, that encloses a set of $m$ individuals, $E$, each being the highest-performing solution of its corresponding niche in behavior space (the elites): 
\begin{flalign*}
& E = \{ \argmax_{\mathbf{x}_1} f(\mathbf{x}_1), \ldots , \argmax_{\mathbf{x}_m} f(\mathbf{x}_m) \} \subseteq H \quad \text{s.t.} \quad \mathbf{x}_i \in C_i
\end{flalign*}
where $C_i \subset X$ is the subset of the genotype space that corresponds to the $i$th region in behavior space, $i = 1 \dots m$, $m \leq k$ and $k$ is the niche capacity.
The objective of an illumination algorithm is to obtain $E$, whose cardinality $m$ typically is in the order of hundreds or thousands.
Finding $H$ from $E$ can be computationally demanding\footnote{We attempted to use the library provided by~\cite{blonder2014n}, but we could not obtain results for the Schwefel (100D) and Hexapod (36D) tasks.}, thus, we would like to use $E$ as a surrogate for $H$ in each task. 

\subsection{Metrics}

\begin{figure}[h]
	\includegraphics[width=0.9\columnwidth]{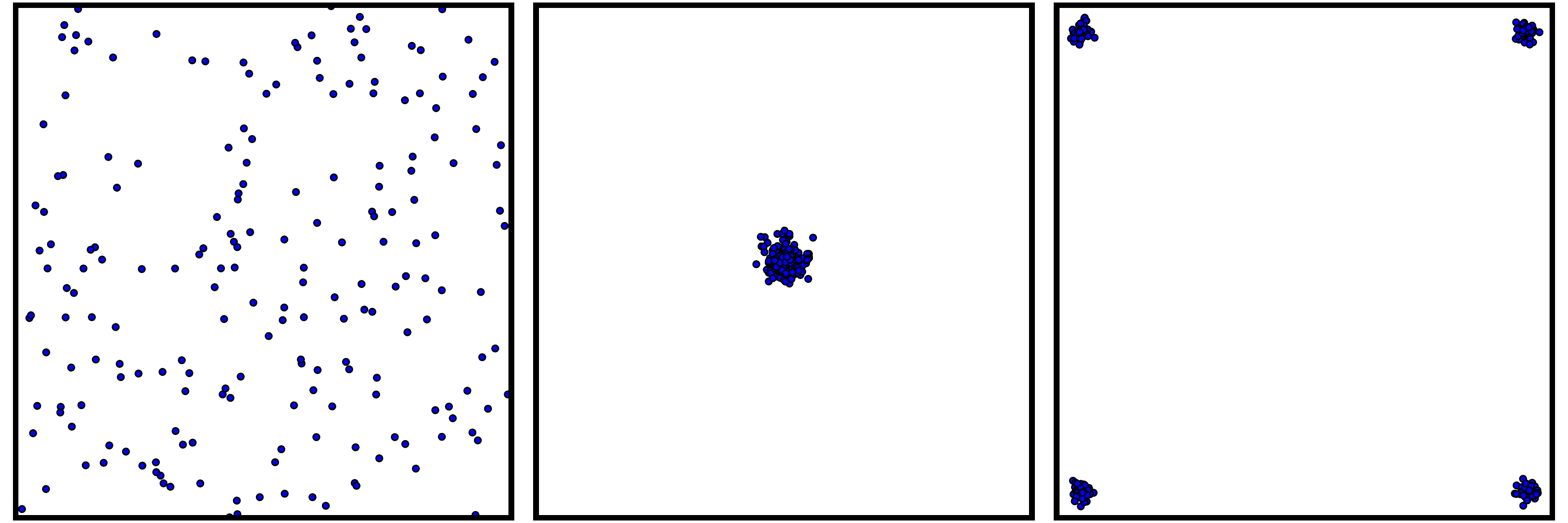}
	\caption{Examples of how the metrics behave. Left: high spread (0.03), low similarity (0.64). Middle: low spread (0.002), high similarity (0.97). Right: low spread (0.002), even lower similarity (0.45).}
	\label{fig_hypervolume_metrics}
\end{figure}
We study each elite hypervolume, $E(t)$, using the archives provided by CVT-MAP-Elites at a given time $t$ (number of evaluations), with the following metrics:
\begin{flalign}
spread(E(t)) = & \frac{ \sum_{\mathbf{x}_i \in E(t)} 
	\min_{\mathbf{x}_j \in E(t), i \neq j} || \mathbf{x}_i - \mathbf{x}_j ||_2 }
{|E(t)| \cdot || \vec{\mathbf{1}} ||_2} \\
similarity(E(t)) = & 1 - \frac{\sum_{\mathbf{x}_i \in E(t)} \sum_{\mathbf{x}_j \in E(t)} || \mathbf{x}_i - \mathbf{x}_j ||_2}{|E(t)|^2 \cdot || \vec{\mathbf{1}} ||_2 }
\end{flalign}

\begin{figure*}
	\includegraphics[width=0.88\textwidth]{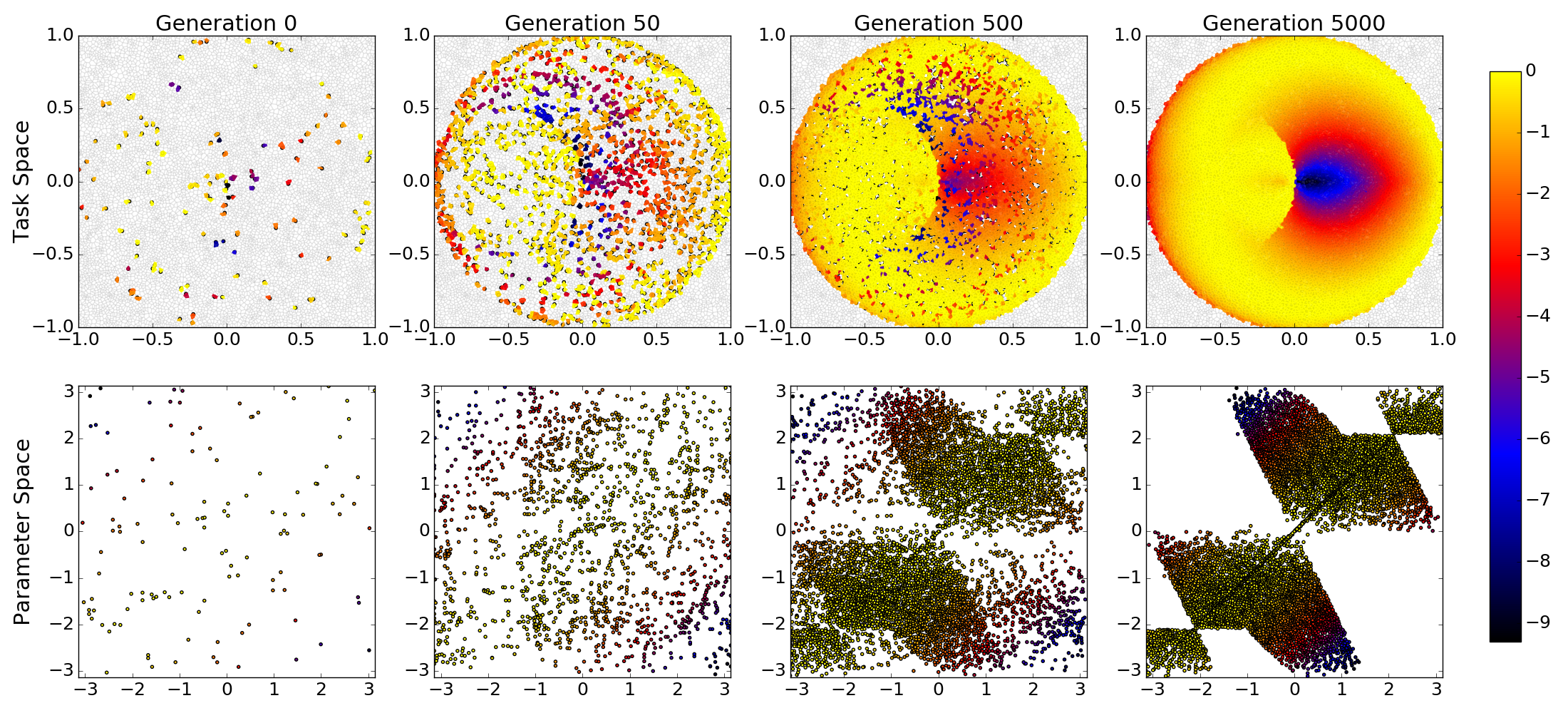}
	\caption{Example with a 2-DOF robotic arm that needs to reach thousands of points in 2D space (upper row; task/behavior space, where niching is performed). Each white region in task space is an empty niche, whereas each colored region corresponds to a genotype, i.e., 2 joint angles (bottom row; genotype/parameter space). The number of points in genotype space show the archive size at a given generation, since only one, elite genotype can occupy a niche.
	The fitness is the negative variance of the joints, thus, solutions along the diagonal (bottom row) have higher fitness. The color reflects the fitness value. The elites of each generation are noisy samples from the volume we are interested in finding (gen. 5000, bottom row).}
	\label{fig_parameter_task_spaces}
\end{figure*} 
Spread can be interpreted as the mean distance to the nearest neighbor normalized by the maximum possible distance (in the bounding volume of the genotype space $[0,1]^n$), and similarity is the mean of the average pairwise distances normalized by the maximum possible distance, thus, representing a fraction, and subtracted from 1 so that a higher value means more similar in terms of percentage.

When taken together, these two metrics\footnote{Other metrics can be considered, such as number of clusters, correlation of solutions in each cluster, volume, geometric span, or others based on manifold learning algorithms, however, the ones we present here are representative for our objective in this paper.} can roughly characterize three different situations: a uniformly spaced set of points has high spread and low similarity (Fig.~\ref{fig_hypervolume_metrics}, left); a single cluster of points has low spread and high similarity (Fig.~\ref{fig_hypervolume_metrics}, middle); and multiple clusters of points have low spread and low similarity (Fig.~\ref{fig_hypervolume_metrics}, right).

\subsection{Elite Hypervolume for each task} \label{sec_hypervolume_tasks}

We run the CVT-MAP-Elites algorithm in each task to study the corresponding elite hypervolume. Any other illumination/quality diversity algorithm could be used instead of CVT-MAP-Elites.

We first run the kinematic arm task with a 2-DOF arm (Fig. \ref{fig_parameter_task_spaces}), which allows us to visualize the elite hypervolume in 2D. The initial, randomly generated elites are evenly spread in the genotypic space (generation 0); however, once CVT-MAP-Elites has converged (here after 5000 generations), the elites are concentrated in a very particular (non-convex) volume in the genotype space (Fig. \ref{fig_parameter_task_spaces}, last panel). The highest performing solutions are on a diagonal line (they correspond to a fitness of $0$, for which the two joint angles are equal), but occupying the other behavioral niches requires to have a lower fitness.
\begin{figure}
	\includegraphics[width=\columnwidth]{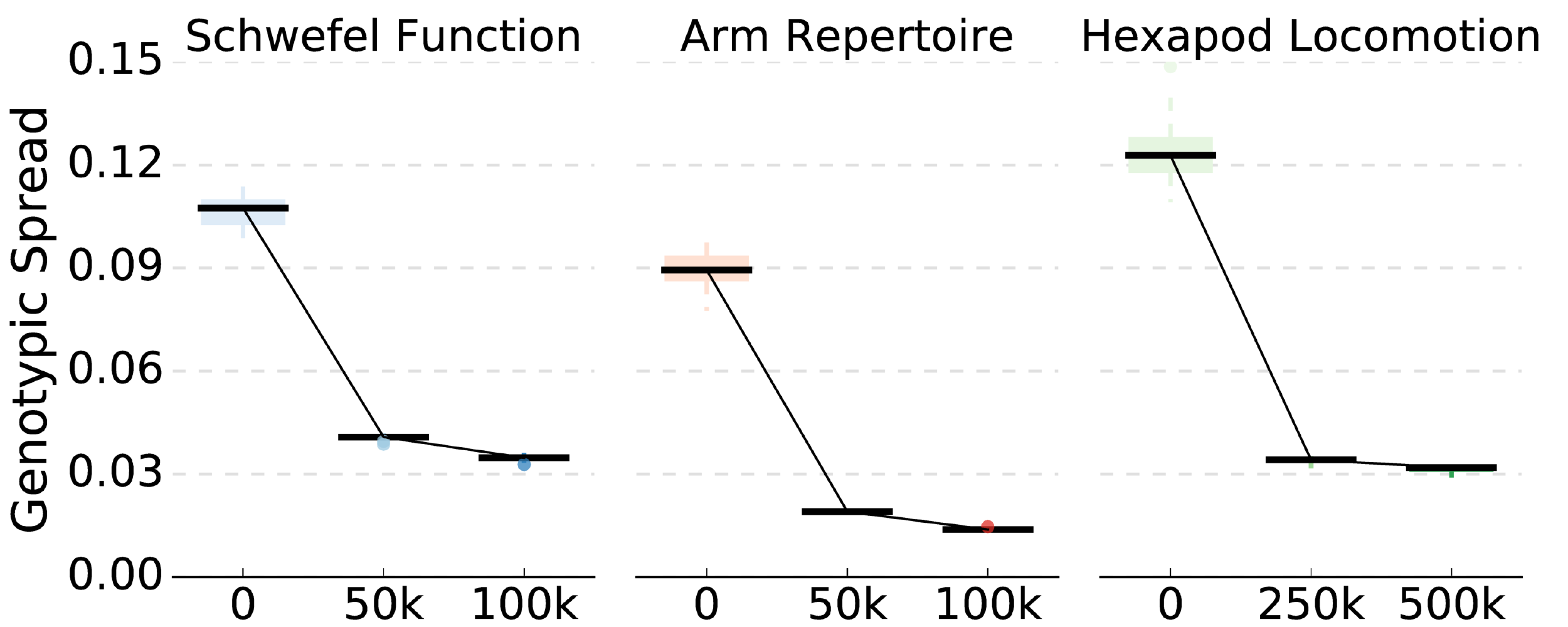}
	\caption{The spread of the elites decreases with the number of evaluations in all tasks.}
	\label{fig_spread}
\end{figure}
\begin{figure}
	\includegraphics[width=\columnwidth]{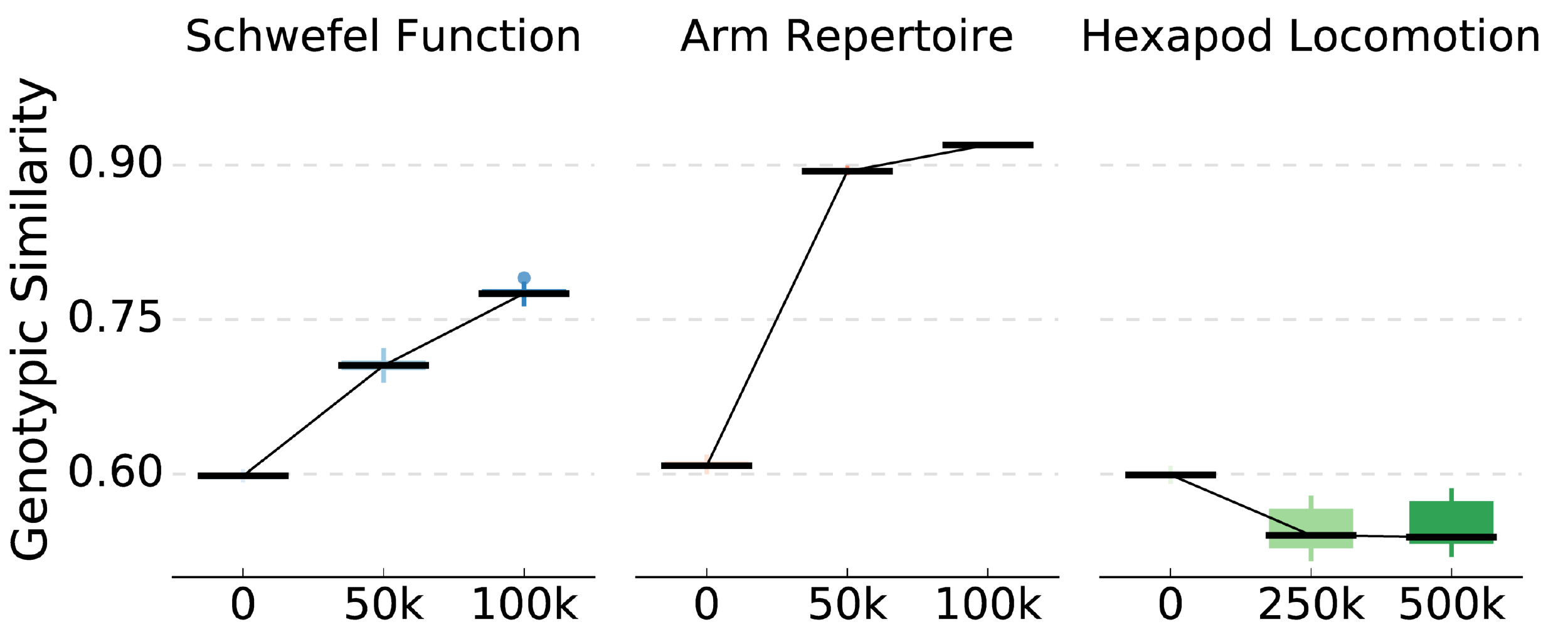}
	\caption{The similarity between elites increases with the number of evaluations in the Schwefel function and arm task, but not in the hexapod task.}
	\label{fig_similarity}
\end{figure}

We then move on with the tasks described in Section~\ref{sec_tasks}. For the experiments with the Schwefel function and the robotic arm, we use 30 replicates of 100k evaluations. The hexapod experiment is a more difficult task, therefore, we use 500k evaluations; since it is computationally more expensive, we use 20 replicates. 

In all tasks, the spread of the solutions becomes lower as we increase the number of evaluations (Fig.\ref{fig_spread}; the differences between pairs are highly significant, $p<10^{-7}$ Mann-Whitney U test). 
On the other hand, the similarity of the elites (Fig.\ref{fig_similarity}) increases with the number of evaluations in the Schwefel function and the arm task ($p<10^{-11}$), but not in the hexapod task ($p=0.64$ between 250k and 500k, $p<10^{-7}$ for the other pairs). This shows that in the Schwefel function and the arm task, the elites become more concentrated into an elite hypervolume, thus, it should be possible to easily exploit their similarities and accelerate illumination. In the hexapod task, however, we should not expect to be able to do so, as the solutions seem to be split into several hypervolumes (clusters).

\section{Directional Variation}
\subsection{Principle and motivation}
When elites share a large part of their genome (here in the first two tasks), it becomes possible to bias the variation operator (the mutation) to make it more likely to generate new candidates in the elite hypervolume. To exploit these inter-species similarities, a simple approach is to extract genotypic correlations and sample new candidates accordingly. In evolutionary computation, this is typically achieved by sampling from a multivariate Gaussian distribution $\mathcal{N}(\mathbf{\mu}, \mathbf{\Sigma})$, where the covariance matrix $\mathbf{\Sigma}$ models the correlations~\cite{hansen2001completely}.

A first idea is to estimate the distribution of all the elites at each generation, that is, to attempt to capture ``global correlations''. However, if we except purely artificial tasks, it is unlikely that all the elites follow the same correlation (this would mean that all the elites lie on a single line in the genotypic space). An alternative is to estimate a multivariate Gaussian distribution from a neighborhood around a parent $\mathbf{x}_i^{(t)}$ (i.e., the objective vector of the elite stored in the $i$th niche), thus, centering the distribution on $\mathbf{x}_i^{(t)}$. Nevertheless, this approach would require to select the appropriate neighborhood, which is likely to be specific to the task and the generation number.

We propose a third approach that exploits the existence of a hypervolume without constructing it, and which is inspired by the success of crossover in extracting the common features of successful individuals (see Sec.~\ref{sec:review-cross}). Once a parent is selected (here we select uniformly among the elites), it is mutated according to the following principles:
\begin{enumerate}
	\item the direction of correlation $\mathbf{d}^{(t)}_{ji}$ is defined by a randomly chosen elite $\mathbf{x}_j^{(t)}$: $\mathbf{d}^{(t)}_{ji} = (\mathbf{x}_j^{(t)} - \mathbf{x}_i^{(t)}) / || \mathbf{x}_j^{(t)} - \mathbf{x}_i^{(t)} ||$;
	
	\item the variance along $\mathbf{d}^{(t)}_{ji}$ depends on the distance $||\mathbf{d}^{(t)}_{ji}||$, so that the mutation is self-adjusting (when the volume shrinks, the variance decreases); this follows from the literature on crossover in real-valued GAs~\cite{deb2001self};
	
	\item to mitigate premature convergence, exploration is performed not only along $\mathbf{d}^{(t)}_{ji}$, but also in all other directions;
	
	\item when $||\mathbf{d}^{(t)}_{ji}||$ is small, the variance does not decrease to zero, to ensure continual exploration in the spirit of illumination algorithms which are inherently exploratory.
\end{enumerate}

\begin{figure}
	\includegraphics[width=\columnwidth]{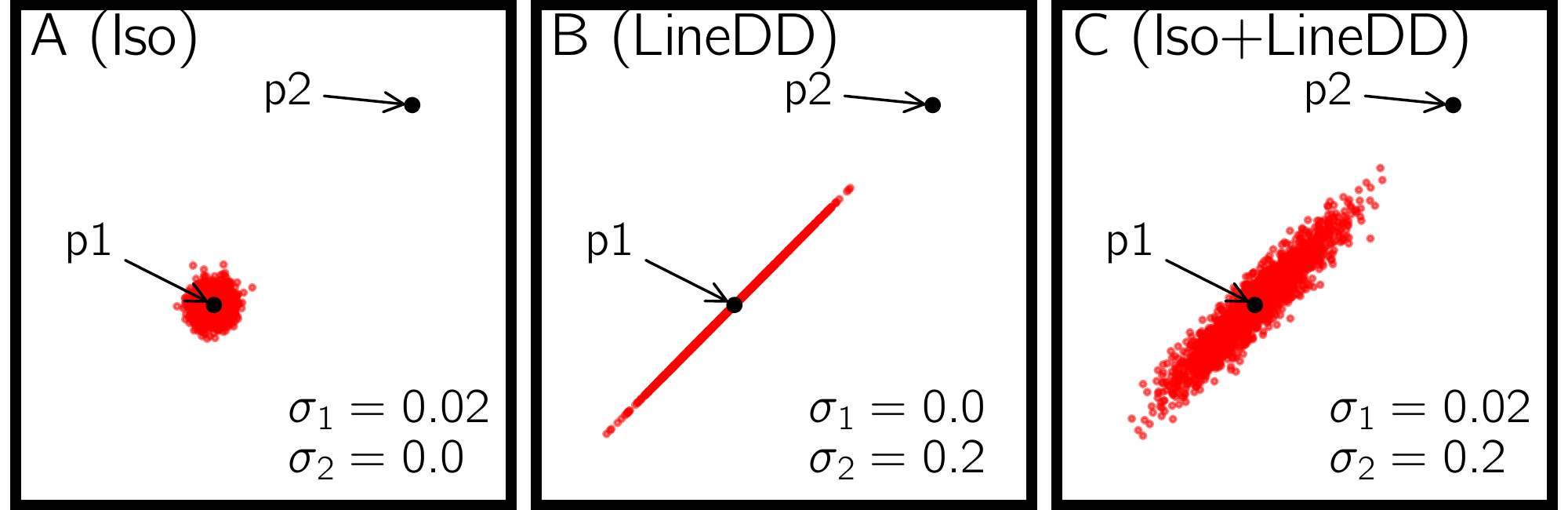}
	\caption{We aim to bias the mutation of an elite $p1$ along the direction of correlation with another elite $p2$. (A) Sampling from an isotropic Gaussian distribution (Iso) centered at $p1$ does not capture any correlations between the 2 points. (B) On the other hand, sampling along the direction of correlation between $p1$ and $p2$ with a distance-dependent variance (LineDD) does not explore all directions around $p1$. (C) Sampling from a multivariate Gaussian distribution centered at $p1$ and pointing at $p2$ (Iso+LineDD) both explores around $p1$ and exploits the correlations between $p1$ and $p2$.}
	\label{fig_directed_mutation}
\end{figure}

We can implement these four principles by convolving two Gaussian distributions (since the result is still a Gaussian distribution).
The first distribution is an isotropic one, which has small variance in all directions (satisfying principles (3) and (4)); the second distribution is a directional one, which adds a Gaussian elongation (satisfying principles (1) and (2)). Under this operator, the offspring is sampled as follows:
\begin{equation}
\label{eq:variation}
\mathbf{x}_i^{(t+1)} = \mathbf{x}_i^{(t)} + \sigma_1 \mathcal{N}(\mathbf{0}, \mathbf{I}) + \sigma_2 (\mathbf{x}_j^{(t)} - \mathbf{x}_i^{(t)}) \mathcal{N}(0, 1)
\end{equation}
where $\sigma_1$ and $\sigma_2$ are user-defined parameters. Note that when $\sigma_2=0$ the effect of the directional distribution disappears and the distribution becomes isotropic (Fig.~\ref{fig_directed_mutation}A). Conversely, when $\sigma_1=0$ the effect of the isotropic distribution disappears and the distribution takes a directional form (Fig.~\ref{fig_directed_mutation}B). When $\sigma_1>0$ and $\sigma_2>0$ the distribution becomes correlated (Fig.~\ref{fig_directed_mutation}C), and as $||\mathbf{d}^{(t)}_{ji}||$ goes to zero, the distribution becomes more and more isotropic. 
Alternatively, Eq.~\ref{eq:variation} can be interpreted as the combination of an isotropic Gaussian mutation and a mutation similar to differential evolution whose scaling factor follows a Gaussian distribution.

\subsection{Evaluation}

In all our experiments, we use the CVT-MAP-Elites algorithm~\cite{vassiliades2017cvt} with 10000 niches. We suspect that similar results would be obtained with other quality-diversity algorithms and variants of MAP-Elites \cite{pugh2016quality,cully2017quality,vassiliades2017unbounded} because our main assumption is that the elites are located in a specific elite hypervolume (and not evenly spread), which does not depend on the way niches are defined.

In addition, we select $\mathbf{x}_i^{(t)}$ and $\mathbf{x}_j^{(t)}$ uniformly at random. It is, however, very likely that directional variation can be combined with biases for novelty or curiosity when selecting parents \cite{cully2017quality,mouret2015illuminating}. Since we care about illumination and not merely finding a single optimal solution, we define three performance metrics, which should all be maximized: (1) archive size (with a maximum capacity of 10000) \cite{mouret2015illuminating}, which corresponds to the number of filled niches, (2) the mean fitness of the solutions that exist in the archive, and (3) the maximum fitness found.

In our experiments, we abbreviate our new variation operator as ``Iso+LineDD'', as it incorporates an isotropic Gaussian (Iso) with a fixed variance and a directional Gaussian (Line) with a distance-dependent (DD) variance. It corresponds to the case where $\sigma_1>0$ and $\sigma_2>0$. In our experiments, we set $\sigma_1=0.01$ and $\sigma_2=0.2$. We evaluate it against the baselines of the next section\footnote{All the parameters settings are found after preliminary experimentation.}.

\pagebreak
\subsection{Baselines}

\subsubsection{Gaussian Line with Distance Dependent Variance (LineDD)} \label{sec_gaussian_line}
This variant of our operator corresponds to the case where $\sigma_1=0$ and $\sigma_2=0.2$ (Fig.~\ref{fig_directed_mutation}B).

\subsubsection{Gaussian Line (Line)}
This simpler variant does not consider a DD variance, thus, producing the offspring as follows: $\mathbf{x}_i^{(t+1)} = \mathbf{x}_i^{(t)} + \sigma \mathbf{d}^{(t)}_{ji} \mathcal{N}(0, 1)$, where $\sigma=0.2$.

\subsubsection{Isotropic Gaussian (Iso)}
This is the standard isotropic Gaussian mutation (Fig.~\ref{fig_directed_mutation}A), for which we set $\sigma=0.1$. 

\subsubsection{Iso with Distance Dependent Variance (IsoDD)}
This variant adapts the variance of the isotropic distribution according to the distance between the two parents:
$\mathbf{x}_i^{(t+1)} = \mathbf{x}_i^{(t)} + \sigma ||\mathbf{x}_j^{(t)} - \mathbf{x}_i^{(t)}|| \mathcal{N}(\mathbf{0}, \mathbf{I})$, where $\sigma=0.05$.

\subsubsection{Iso Self-Adaptation (IsoSA)}
Our new operator uses a DD variance along the direction of correlation between elites. This can be seen as a type of self-adaptation~\cite{deb2001self}, thus, it is natural to ask whether the type of self-adaptation found in ES~\cite{beyer2002evolution} confers any similar benefits. We thus extend each individual to additionally contain a single mutation strength $\sigma$ that is updated using a log-normal distribution: $\sigma_i^{(t+1)} = \sigma_i^{(t)} \, exp(\tau \, \mathcal{N}(0, 1))$, where $\tau = (2n)^{-1/2}$ and $\sigma_i^{(0)} = 0.1$. The offspring is then generated as follows: $\mathbf{x}_i^{(t+1)} = \mathbf{x}_i^{(t)} + \sigma_i^{(t+1)} \, \mathcal{N}(\mathbf{0}, \mathbf{I})$.

\subsubsection{Global Correlation (GC)}
At every generation, we estimate the direction of global correlation by fitting a multivariate Gaussian distribution on all the points of the archive, i.e., $\mathcal{N}(\mu_{global}, \Sigma_{global})$. Then we sample the offspring for each selected parent, by centering the distribution on the parent and scaling the covariance by some factor $\alpha$: $\mathbf{x}_i^{(t+1)} = \mathbf{x}_i^{(t)} + \alpha \, \mathcal{N}(\mathbf{0}, \Sigma_{global})$\footnote{Our experiments have shown that it is \textit{always} better, not to sample from the mean of the distribution, but to use this parent-centric approach.}. We set $\alpha=0.1$.

\subsubsection{Simulated Binary Crossover with One Offspring (SBX)}

Since our variation operator resembles a crossover operator for real-coded GAs, we compare it with SBX~\cite{agrawal1995sbx}, which is also parent-centric, and self-adjusting according to the distance between the two parents. To have a fairer comparison, instead of creating two offspring (each one near its corresponding parent), we generate only one as follows: $x_{ik}^{(t+1)} = 0.5 \big[ (1+\beta_k) \, x_{ik}^{(t)} + (1-\beta_k) \, x_{jk}^{(t)} \big]$,
where $x_{ik}^{(t)}$ is the $k$th element of solution $\mathbf{x}_i^{(t)}$ ($k = 1, 2, \dots, n$), and
\begin{equation*}
\beta_k =
\left\{
\begin{array}{ll}
(2 \, u_k)^\frac{1}{\eta + 1}  , & \mbox{if } u_k \leq 0.5; \\
\bigg(\frac{1}{2 \, (1 - u_k)}\bigg)^\frac{1}{\eta + 1} , & \mbox{otherwise.} 
\end{array}
\right.
\end{equation*}
where $\eta \geq 0$ is the distribution index and $u_k$ is a random number generated from a uniform distribution in $[0,1)$. In addition, each variable has a $0.5$ chance of not being recombined in which case $x_{ik}^{(t+1)} = x_{ik}^{(t)}$. In our experiments, we set $\eta = 10$.

\subsection{Results}

\begin{figure*}
	\includegraphics[width=0.85\textwidth]{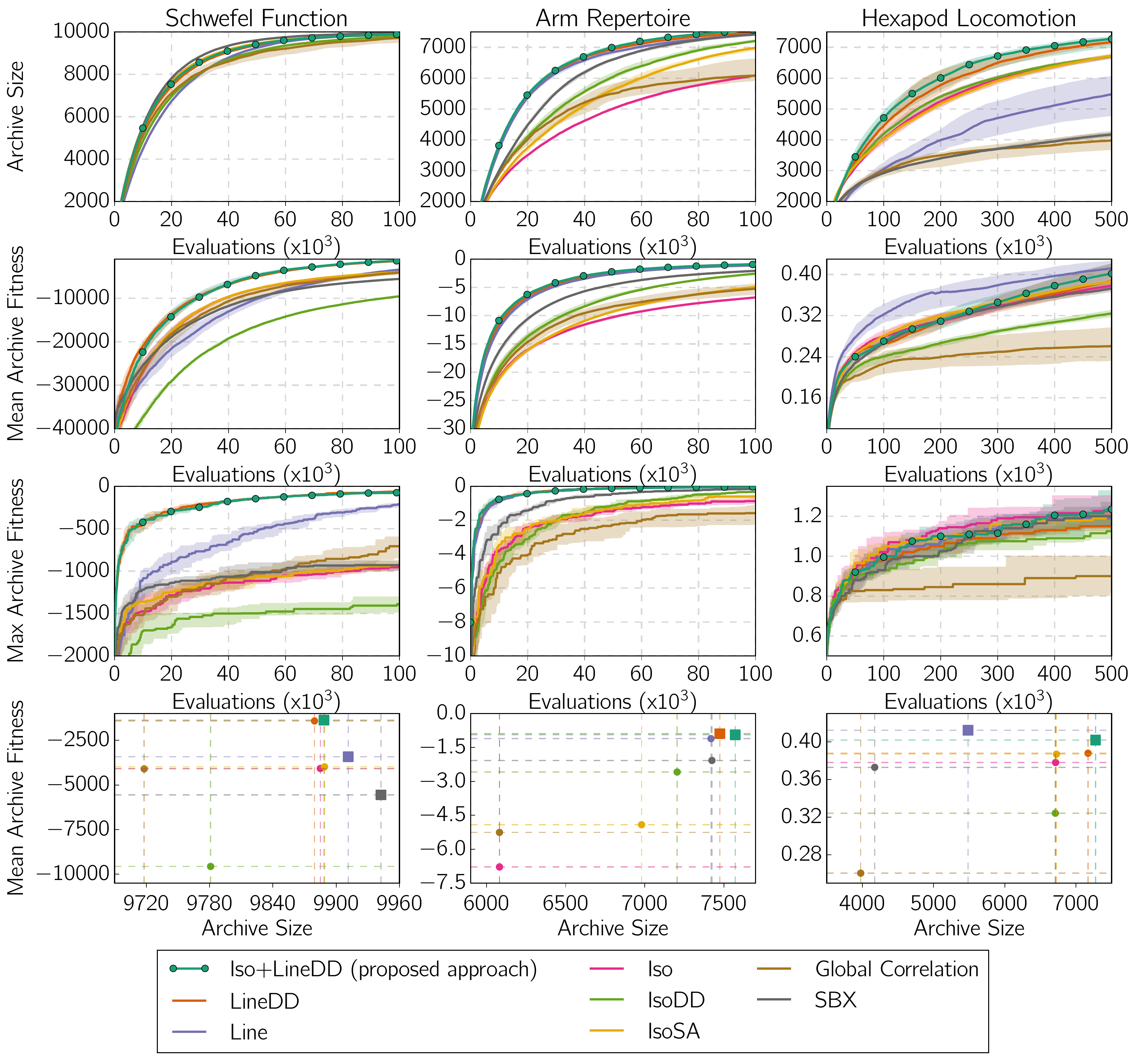}
	\caption{Results of different variation operators. Iso+LineDD, LineDD, and Line, have significant advantages in terms of: performance in the Schwefel function (left column); diversity and performance in the arm task (middle column); diversity in the hexapod task (right column). The plots show the median and the interquartile range ($25^{th}$ and $75^{th}$ percentiles) over 30 replicates for the Schwefel function and the arm task, and 20 replicates for the hexapod task. The bottom row illustrates the final solutions in terms of the median of the mean fitness against the median of the archive size: non-dominated operators are shown as squares, whereas dominated ones are shown as circles. Iso+LineDD is never Pareto dominated by any other operator.}
	\label{fig_results}
\end{figure*}

As in Sec.~\ref{sec_tasks}, we use 30 replicates of 100k evaluations for the Schwefel function and the arm task and 20 replicates of 500k evaluations for the hexapod task. All the reported results are medians over these runs.
The results (Fig.~\ref{fig_results}) confirm our conclusions when measuring the spread (Fig.~\ref{fig_spread}) and similarity (Fig.\ref{fig_similarity}) of the corresponding elite hypervolume (Sec.~\ref{sec_hypervolume_tasks}). In particular, Iso+LineDD and LineDD accelerate illumination in Schwefel's function (in terms of mean and max performance; Fig.~\ref{fig_results}, left column, 2nd and 3rd panel) and the arm task (in all metrics; Fig.~\ref{fig_results}, middle column), while for the hexapod task they only provide marginal benefits (slightly better progress in archive size; Fig.~\ref{fig_results}, right column, 1st panel).

More specifically, in Schwefel's function, Iso+LineDD and LineDD, reach a level of maximum performance at 10k evaluations (-416.5 and -419.7 respectively) that is only reached at 60k evaluations with Line (-439.7) and never reached by any other operator at 100k evaluations. This demonstrates an order of magnitude faster improvement. It also shows that DD variance is beneficial when coupled with the line operators, since IsoDD has the worst overall performance. The archive size increases at the same rate with all operators (1st panel); however, this is expected since the behavior space is just a subset (rather than a function) of the genotype space. 

In the arm task, all line operators have the best progress rates in all 3 metrics, followed by SBX. Self-adaptation (SA) helps the Iso operator in attaining better progress rates for archive size and mean fitness, and DD helps to accelerate its progress even more in contrast to the previous task. While GC always surpasses Iso in terms of mean fitness, Iso has better overall max fitness and reaches the archive size of GC (6082) at 100k evaluations.

Finally, in the hexapod task, Iso+LineDD and LineDD have the lead in terms of archive size (7274 and 7169 at 500k evaluations respectively), significantly outperforming the Iso operators which come second (6710-6725; $p<10^{-4}$ Mann Whitney U test); Line does not perform as well (5483.5) and displays high variance, while SBX and GC come last (4173.5 and 3977 respectively). However, in terms of mean fitness, the Line operator starts off faster than the other operators, however, at 500k evaluations Iso+LineDD manages to reach a similar level, with their difference \textit{not} being statistically significant (Line: 0.412, Iso+LineDD: 0.402, $p=0.18$ Mann-Whitney U test). Although this shows that making smaller steps along the direction of correlation helps in improving the solutions in Line's archive, note that the archive size of Line at 500k evaluations (5483.5) is even lower than the one of Iso+LineDD at 150k evaluations (5502). SA does not significantly affect the performance of Iso, whereas, DD negatively impacts it both in terms of mean and max fitness. In all three metrics, GC is consistently worse than the other operators.

When plotting the final solutions found by the operators in terms of median of the mean fitness against the median of the archive size (Pareto plot; Fig.~\ref{fig_results}, bottom row), we observe that Iso+LineDD is never Pareto dominated in all three tasks.

\section{Conclusion and Discussion}

In this paper, we demonstrated that when using illumination algorithms in certain tasks (here the Schwefel function and the arm experiment), the set of solutions returned by the algorithms form an elite hypervolume in genotype space. We introduced two metrics, the genotypic spread and the genotypic similarity, to empirically characterize this hypervolume, as well as a variation operator that can exploit correlations between solutions. We then showed that in case the elite solutions display high genotypic similarity, the operator can significantly increase the progress rate of MAP-Elites in terms of performance (without reducing diversity, e.g., in the Schwefel function) or both performance and diversity (e.g., in the arm task); in case the elite solutions display low genotypic similarity, the operator can significantly increase the diversity (without reducing performance, e.g., in the hexapod task).

The variation operator we introduced here resembles a parent-centric crossover for real-coded GAs. It plays a significant role in illumination algorithms because it provides them with a better balance between exploration and exploitation. In other words, the niching scheme provides the exploration, while this operator provides the exploitation because it has the ``right'' bias. Thus, we expect it to be more useful in cases where there is a good diversity of solutions. To our knowledge, such results are lacking in the field of real-coded GAs.

It is likely that selective biases for illumination algorithms~\cite{pugh2016quality, cully2017quality}, will complement the variation biases we introduced here, thus, further accelerating illumination. For instance, we could minimize the chances of sampling the regions outside the elite hypervolume by restricting the selection of the \textit{second} elite which would define the direction of correlation. Such an approach bears similarities to the restricted tournament selection~\cite{harik1995rts} method from multimodal optimization. Taking more inspiration from the multimodal optimization literature~\cite{vassiliades2017multimodal}, we could select the second elite to be the nearest \textit{better} neighbor~\cite{preuss2005counteracting} of the first, in which case we would bias for performance.

One might wonder whether the findings of this work apply for variable-sized genotypes, such as the ones used by the NeuroEvolution of Augmenting Topologies (NEAT) algorithm. While the crossover operator of NEAT is effective in recombining variable-sized neural networks, or compositional pattern producing networks~\cite{stanley2009hypercube}, it resembles a disruptive, mean-centric approach to recombination (e.g., see~\cite{eshelman1993blx, ono1997undx, tsutsui1999spx}), rather than a parent-centric one (e.g., \cite{agrawal1995sbx, deb2002computationally}). Thus, an interesting research direction would be to study how correlations between graphs can be modeled and exploited. Model-building approaches for genetic programming~\cite{kim2014probabilistic} could provide a fruitful inspiration for such an endeavor.

\begin{acks}
	This work was supported by the European Research Council under the European Union's Horizon 2020 research and innovation programme (Project: ResiBots, grant agreement No 637972). We would like to thank Francis Colas and Konstantinos Chatzilygeroudis.
\end{acks}

\appendix

\section*{Source Code}
The source code of the experiments can be found in {\footnotesize\url{https://github.com/resibots/vassiliades_2018_gecco}}.

\bibliographystyle{ACM-Reference-Format}
\bibliography{refs} 

\end{document}